\documentclass[final]{cvpr}
\usepackage{times}
\usepackage{epsfig}
\usepackage{graphicx}
\usepackage{amsmath}
\usepackage{amssymb}
\usepackage{multirow}
\usepackage[page,toc,titletoc,title]{appendix}
\usepackage{titletoc}
\usepackage{booktabs}
\usepackage{bm}
\usepackage{dsfont}
\usepackage{subfigure}
\usepackage{pifont}
\usepackage{parskip}
\usepackage[nocompress]{cite}
\usepackage[pagebackref=true,breaklinks=true,colorlinks,bookmarks=false]{hyperref}
\usepackage{cleveref}
\usepackage{textcomp}
\usepackage[numbers,sort]{natbib} 
\usepackage{pifont}
\usepackage[table,xcdraw]{xcolor}
\usepackage{longtable}
\usepackage{pdflscape}
\usepackage{caption}
\usepackage{wrapfig}
\usepackage{enumitem}

\usepackage[pagebackref=true,breaklinks=true,colorlinks,bookmarks=false]{hyperref}

\def\confYear{CVPR 2021}

\newcommand{\cmark}{\ding{51}}%
\newcommand{\xmark}{\ding{55}}%

\renewcommand{\paragraph}[1]{\noindent{\bf #1}\quad}

\newif\ifarxiv

\arxivtrue  

\begin{document}
	
	\title{Hand-Object Interaction Reasoning}
	
	\author{Jian Ma\\
		University of Bristol\\
		{\tt\small jian.ma@bristol.ac.uk}	
		\and
		Dima Damen\\
		University of Bristol\\
		{\tt\small dima.damen@bristol.ac.uk}}
	
	\maketitle

	\urlstyle{same}

	\begin{abstract}
		This paper proposes an interaction reasoning network for modelling spatio-temporal relationships between hands and objects in video. The proposed interaction unit utilises a Transformer module to reason about each acting hand, and its spatio-temporal relation to the other hand as well as objects being interacted with. We show that modelling two-handed interactions are critical for action recognition in egocentric video, and demonstrate that by using positionally-encoded trajectories, the network can better recognise observed interactions. We evaluate our proposal on EPIC-KITCHENS and Something-Else datasets, with an ablation study.
	\end{abstract}

	\section{Introduction}
\label{sec:intro}

Different from general actions (e.g. jumping), object interactions involve actors influencing objects (e.g. playing an instrument or kicking a ball).
Of particular interest to this work is hand-object interactions (HOI) which feature regularly in the activities of daily living.
HOIs include one-handed (e.g. ``open drawer'') as well as two-handed interactions (e.g.~``open bottle''), and many interactions include tools that extend our hands' abilities (e.g. cutting a vegetable requires a knife). 
However, most video understanding methods aim to recognise both actions and interactions alike
as general video datasets involve a mix of classes~\cite{carreira2017quo, Gu_2018_CVPR}.
Recently, a handful of large-scale datasets~\cite{goyal2017something,  Damen2018EPICKITCHENS,Damen2020RESCALING} that focus on HOIs have fueled works that specifically reason about interactions. 

Recent progresses in interaction reasoning have been driven by the success of object detectors, e.g. \cite{ren2015faster}.
Due to the datasets used, previous works~\cite{sun2018actor, girdhar2019video, pan2020actor, Baradel_2018_ECCV} detect the person in the middle frame of the video or all the objects that appear in the video. This means that trajectories of interactions are not explicitly emphasised.

\begin{figure}[t]
\centering
\includegraphics[width=1\linewidth]{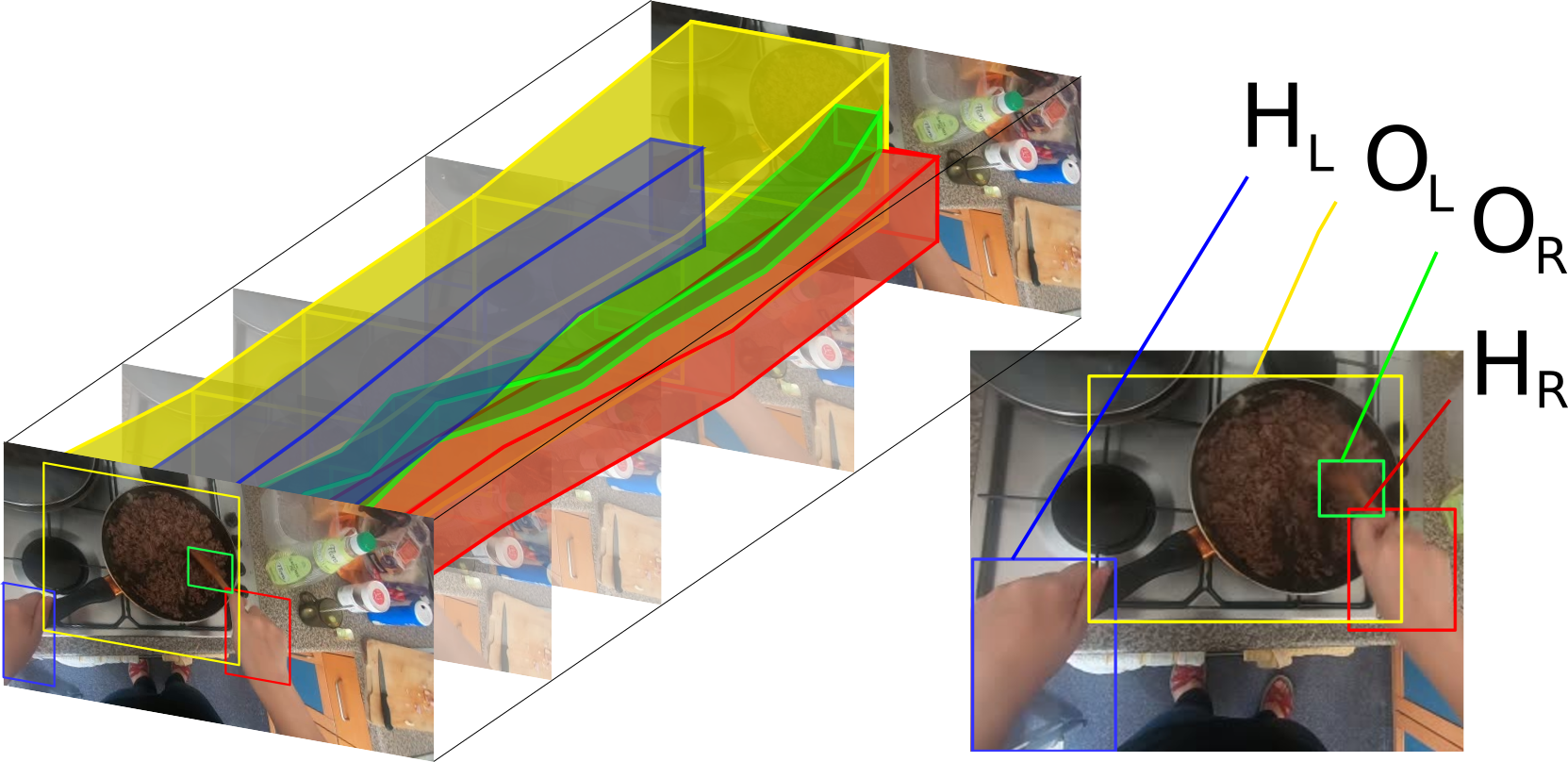}
\caption{Illustration of Spatio-Temporal HOI reasoning for the action of stirring food in the pan. Both hands play critical roles. The left hand ($H_{L}$, in blue) steadies the pan, which we refer to as the left object ($O_{L}$, in yellow) given its direct interaction with the left hand. The right hand ($H_{R}$, in red) holds the wooden spoon ($O_{R}$, in green). While $H_L$, $O_L$ are steady over time, the red and green trajectories ($H_R, O_R$) demonstrate the stirring motion relative to the pan.}
\vspace*{-1pt}
\label{fig: intro}
\end{figure}


In this work, we focus on the interaction between hands and objects, through their trajectories that encode motion and positions, thus discriminating between different interactions~(Fig.~\ref{fig: intro}). 
We propose an \textbf{I}nteraction \textbf{R}easoning \textbf{N}etwork~(IRN) that jointly reasons about interactions between both hands and active objects. 

We propose encoders that learn hand-object and hand-hand interactions and decoders that enrich this learned interaction with action representation knowledge. We automatically detect hands and active objects, i.e. those with which the hand interacts, using the approach from~\cite{Shan20}, and link these detections over time to form trajectories with pooled features from a spatio-temporal backbone.
We then reason about pairwise interactions, distinguishing left from right hand interactions. The proposed framework is trained end to end.

\begin{figure*}[t]
\centering
\includegraphics[width=1\linewidth]{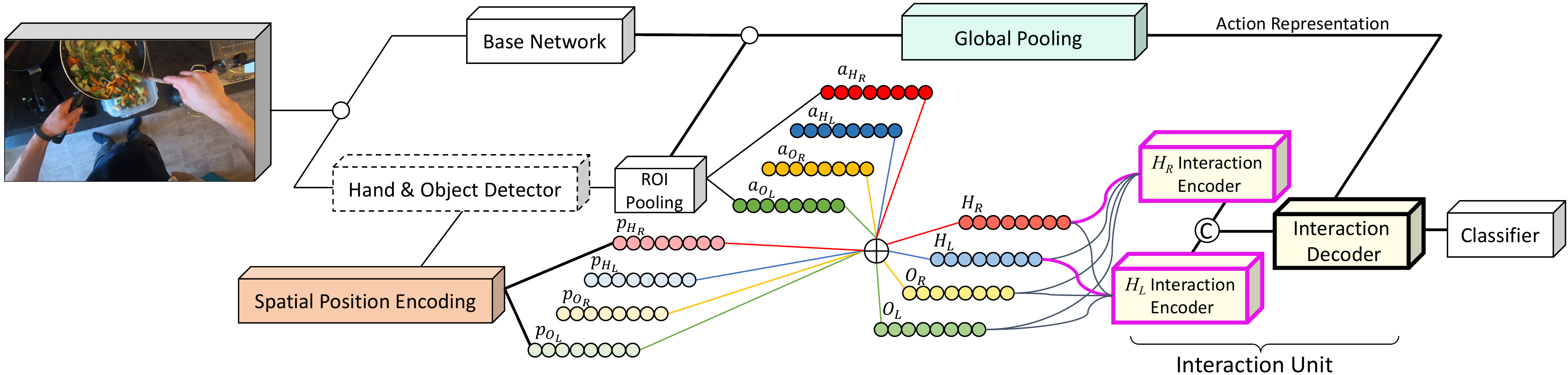}
\vspace*{-10pt}
\caption{Proposed Architecture for Interaction Reasoning Network (IRN). A base 3D ConvNet extracts temporal-spatial features. These are ROI pooled based on hand-object detections and combined with positional encoding to form trajectories. The trajectories are fed into interaction unit to reason about hand-object relations. A decoder combines the action representation and two-hand encoders for classification (Fig.~\ref{fig: IU} for details) . $\circ$ is for multiple outputs, $\oplus$ is for summation and \emph{C} for concatenation. Dashed box for `Hand \& Object Detector' highlights frozen weights. Thicker lines highlight backpropagation pathways. }
\vspace*{-12pt}
\label{fig: overall}
\end{figure*}

\noindent \textbf{Our contributions}: (i) this paper proposes to separately reason about interactions between the left and right hands with their corresponding objects in HOI through an encoder-decoder transformer module; (ii) to the best of the authors' knowledge, the proposed method is the first work that uses the trajectories of hands and objects to enrich the relational representation of interactions; (iii) we showcase the importance of this proposed reasoning on the large-scale egocentric HOI dataset EPIC-KITCHENS-100~\cite{Damen2020RESCALING} as well as on the crowd-sourced HOI dataset \mbox{Something-Else}~\cite{materzynska2020something}.




\section{Related Work}
\label{sec: related_work}

\noindent {\bf Action Recognition.} In recent years, the success of deep learning in computer vision has promoted the rapid development of action recognition models. From the 2D ConvNets~\cite{yue2015beyond, donahue2015long, wang2016temporal, zhou2018temporal, li2019collaborative} and multi-stream networks~\cite{simonyan2014two, feichtenhofer2016convolutional, carreira2017quo} to 3D ConvNets \cite{ji20123d, tran2015learning, feichtenhofer2019slowfast, Feichtenhofer_2020_CVPR} and transformer-based networks~\cite{neimark2021video, Bertasius2021}, these models have progressively improved the understanding of actions. 
Of relevance to our model, SlowFast network \cite{feichtenhofer2019slowfast} is a 3D convolutional model that can combine the spatial features with temporal information from two rates of sampling the input video. SlowFast results remain competitive particularly for HOI datasets we analyse. 
These approaches form the base for interaction reasoning which we review next.

\noindent
{\bf Interaction Recognition.} 
Recently, interest in actor-centric video understanding employed increasingly reliable person detectors, to localise actors in movies~\cite{Gu_2018_CVPR} or players in sports~\cite{soomro2014action}. 
Many works \cite{gkioxari2018detecting, chao2018learning, li2019transferable, wan2019pose, gao2018ican, gao2020drg, pan2020actor, lfb2019, girdhar2019video, materzynska2020something, Shan20, zhou2019relation} utilise an object detector (\emph{e.g.} Faster R-CNN) to locate bounding boxes of human or objects for input into a relational module. However, the detector is likely to yield object proposals that are not relevant to the interaction due to the lack of annotations for training. Instead, \cite{gkioxari2018detecting} learns to predict the location of related objects based on the appearance of actors. 
Similarly,~\cite{chao2018learning, gao2018ican} introduce pairwise streams for interaction patterns to encode the spatial relative locations of human and objects. Based on pairwise streams, human poses are considered for modeling HOIs in \cite{li2019transferable, wan2019pose}. Additionally, Graph Convolutional Network (GCN) is another way to explore image-based object interactions~\cite{zhou2019relation, gao2020drg}.

Different from image-based methods, the motion is also particularly important in video analysis. Given person detections from the middle frame of a video, in \cite{girdhar2019video}, detections are pooled as a query to attend to the whole frame's 3D features, in a transformer encoder block. Longer-term reasoning is proposed in \cite{lfb2019} by learning contextual information through short-term person feature banks and long-term feature banks from non-local blocks \cite{wang2018non}. Apart from the contextual interaction in the temporal dimension,~\cite{Baradel_2018_ECCV} study relational interactions between objects via training a Gate Recurrent Unit with pairing current and previous frames. Similarly,~\cite{wang2018videos} regard video as a graph of objects, conducting interaction recognition reasoning on a graph network. Moreover, \cite{materzynska2020something} generalise the performance to unseen actions, decomposing each action into a verb, subject, and one or more goals and proposing the Something-Else dataset for exploring hand-object interactions.
Detections over time are combined through the Hungarian algorithm to track the detected persons and objects. However, detected objects may not be part of the interaction, which introduces noise in all but the simplest scenes. 
Besides, in~\cite{materzynska2020something}, additional expensive-to-collect annotations of objects are required to train the model.

IRN is inspired by~\cite{Baradel_2018_ECCV,girdhar2019video}, but we focus on hands as actors, thus requiring to model two-hand interactions including hand-to-hand interactions as well as hand-to-object interactions.
Different from~\cite{materzynska2020something}, we do not require additional annotations and instead use automatic detection of hands and active objects from~\cite{Shan20}. This allows us to reason about interactions in busy scenes efficiently. 
We detail our method next.

\section{Methodology}
\label{sec:method}

In this section, we firstly give an overall of the Interaction Reasoning Network (IRN) architecture (visualised in Fig.~\ref{fig: overall}). We then discuss detections and the spatial position encoder followed by details of our proposed interaction unit (IU).


IRN captures the relationships between hands and active (i.e. action-relevant) objects. 
We 
model interactions through i) encoders that focus on hand-object and hand-hand interaction; and ii) decoders that enrich the action representation, trained jointly. 
We formulate the interaction unit as:
\begin{equation}
\label{equ:overall}
\begin{aligned}
E&=e(H, O, P; \omega_{e}), \\
I&=d(F, E; \omega_{d}), \\
\end{aligned}
\end{equation}
where $\omega_{e}$ and $\omega_{d}$ represent the parameters of encoders $e()$ and decoders $d()$. 
The encoder aims to model the hand $H$ and interacting object $O$ representations along with their spatial position encoding $P$.
The encoder's output $E$ and global action representation $F$ are fed into decoders. 

\begin{figure}[t]
\centering
\includegraphics[width=\linewidth]{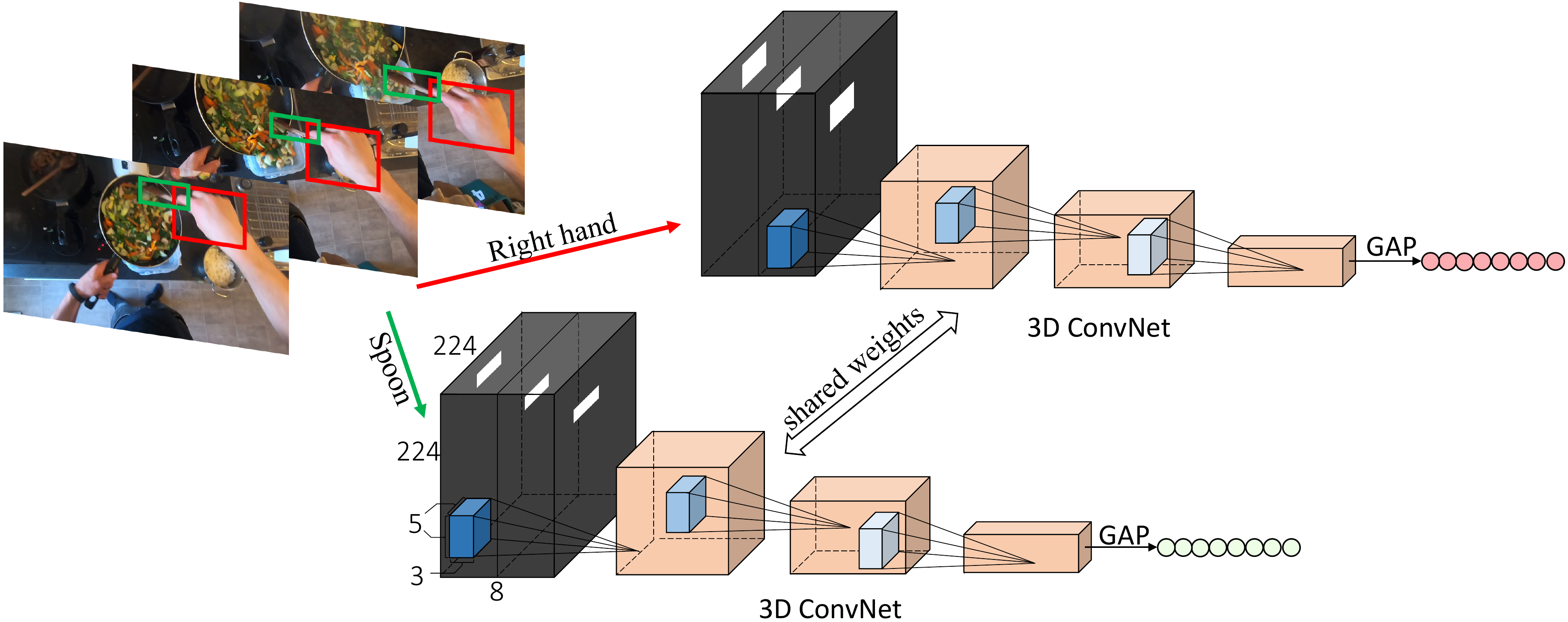}
\caption{Spatial Position Encoder. Detections, per object, over a sequence are represented by binary maps. A 3-layer 3D ConvNet produces per-frame spatial encoding.}
\label{fig: spe}
\vspace*{-1pt}
\end{figure}

\subsection{Detections and Positional Encoding}
\label{subsec: HOD}


We use a pretrained hand and object detector, which we run per frame, and distinguish four bounding box detections of $\left \{ b_{H_{L}}, b_{H_{R}}, b_{O_{L}}, b_{O_{R}} \right \}$ where $H/O$ mean hand and object, $L/R$ denote the side, left or right. 
We pool a 3D ConvNet to extract frame-level hands and objects features, using RoI average pooling as in~\cite{Baradel_2018_ECCV,girdhar2019video}. Given a layer in the backbone with $C$ channels, we extract a feature $a$ per detection of size~$\mathbb{R}^{C}$ by RoI pooling, of which we have $\left \{ a_{H_{L}}, a_{H_{R}}, a_{O_{L}}, a_{O_{R}} \right \}$.


\begin{figure*}[t]
\centering
\includegraphics[width=\linewidth]{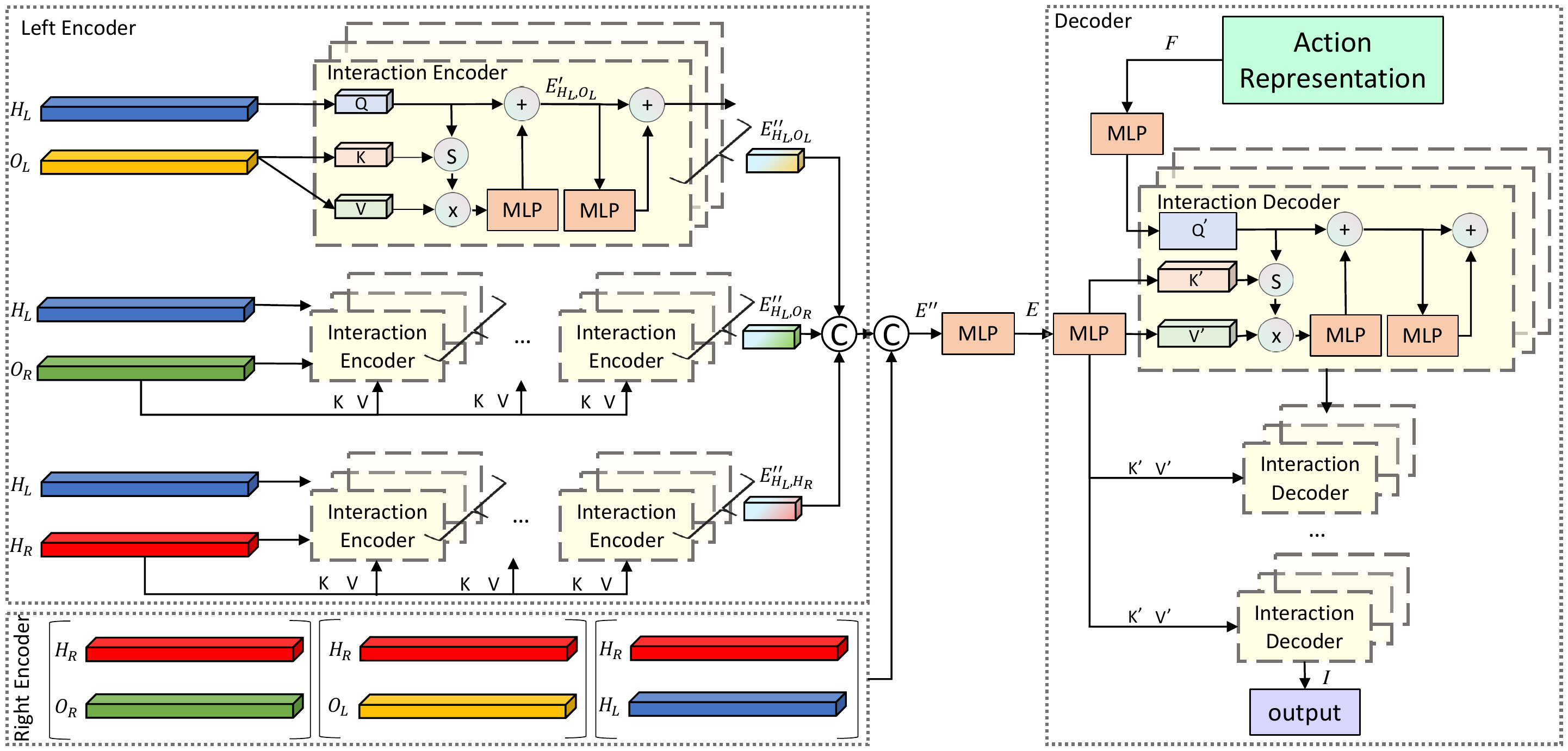}
\caption{The Architecture of proposed Interaction Unit (IU). This module includes encoders and decoders. Encoders reason about hand-object interactions by multi-head transformers with hands as query and objects as key and value. Encoders outputs from left and right hands are concatenated and fed to decoders. Decoders model the interaction with action representation.} 
\vspace*{-12pt}
\label{fig: IU}
\end{figure*}

In addition to the features, the absolute positions of hands and objects offer significant information to distinguish interactions. We thus propose to use spatial positional representation $p$ for each detection, such that $\left \{ p_{H_{L}}, p_{H_{R}}, p_{O_{L}}, p_{O_{R}} \right \}$. Inspired by~\cite{islam2020much}, we consider a binary map for each detection $b$. These show the absolute position and scale of an object in the image. We learn position (and scale) encodings from this binary input and use the consecutive binary maps to capture the positional/scale changes. 
Note that while we use local 3D convolution to model motion, we have the position encoding per frame to correspond to the bounding box $b$.
We learn a weights-shared 3D ConvNet 
as shown in Fig.~\ref{fig: spe}. 

As interactions focus on the temporal evolution of the relationships between hands and objects, we combine features and their positional encodings to form trajectories. We use standard summation but also ablate this against concatenation. We represent each trajectory ${H = (a^1 + p^1, \cdots, a^T + p^T)}$ over $T$ frames, and similarly for objects. 
We thus have four trajectories $\left \{ H_{L}, H_{R}, O_{L}, O_{R} \right \}$, which form the input for our interaction unit.
When hands or objects are not present in a frame or are not detected, we set the feature $a$ to~0 and the binary map to blank.
We showcase experimentally that our method can recover from missed detections.
Additionally, some trajectories might be missing altogether. 
We consider the presence as well as the absence of hands and objects as informative evidence for interaction reasoning.

We next describe how we can reason about interactions using these trajectories.


\subsection{Interaction Unit}
\label{subsec: HOI}



In order to reason about HOIs, we consider a pair of trajectories ($H$, $O$) where $H$ is the actor - a left/right hand and $O$ is the object with which the hand interacts. 
Importantly $O$ can be the other hand or an object.
As Fig.~\ref{fig: IU} demonstrates, we have up to 3 interactions per actor/hand. For the left hand, these would be $(H_L, O_L)$, $(H_L, O_R)$ and $(H_L, H_R)$.
We intuitively describe these using the example from Fig.~\ref{fig: intro} on stirring food in the pan.

\vspace*{-8pt}
\begin{itemize}[leftmargin=*,itemsep=-2ex,partopsep=1ex,parsep=2ex]
    \item $(H_L, O_L)$ captures the left hand holding the pan. The hand gesture and relative positions of hand and pan would be captured using this interaction.
    \item $(H_L, O_R)$ captures the left hand versus the spoon as it stirs through the food in the pan held by the left hand.
    \item $(H_L, H_R)$ captures the absolute positions and gestures of both hands as one holds the pan and the other stirs the food. 
    \end{itemize}
We first describe our encoders.
These are concatenated and passed to stacked decoders described after.

\noindent \textbf{Interaction Encoder.} We explain the encoder for one interaction pair namely $(H_L, O_L)$. 
\begin{equation}
\begin{aligned}
Q_{H_{L}, O_{L}}&=q(H_{L}), \\
K_{H_{L}, O_{L}}&=k(O_{L}), \\
V_{H_{L}, O_{L}}&=v(O_{L}),
\end{aligned}
\end{equation}
where $q, k, v$ all linearly project $\mathbb{R}^{T\times C}$ input to $\mathbb{R}^N$. Fig.~\ref{fig: IU} shows the module of left-hand interactions.
The encoder is then a residual attention unit:

\begin{equation}
E^{'}_{H_{L}, O_{L}}=\sigma\left(\frac{Q_{H_{L}, O_{L}}K_{H_{L}, O_{L}}}{\sqrt{N}}\right)V_{H_{L}, O_{L}}+Q_{H_{L}, O_{L}},
\end{equation}
where $E^{'}_{H_{L}, O_{L}}$ is interaction representation between actor $H_L$ and object $O_L$ and $\sigma$ is the softmax operator. Following~\cite{vaswani2017attention}, we add a linear feedforward network $FFN()$:
\begin{equation}
E^{''}_{H_{L}, O_{L}}=\delta(FFN(E^{'}_{H_{L}, O_{L}}))+E^{'}_{H_{L}, O_{L}},
\label{equ: ffn}
\end{equation}
where $\delta$ is the dropout operation and $E^{''}_{H_{L}, O_{L}}\in \mathbb{R}^{N}$ is the encoded representation for $(H_{L}, O_{L})$. 

Similarly, $E^{''}_{H_{L}, O_{R}}$ and $E^{''}_{H_{L}, H_{R}}$ are computed for the left hand (see Fig.~\ref{fig: IU} Left Encoder) as well as three encoders for $E^{''}_{H_R}$. We concatenate all outputs to form overall encoding ${\mathbf{E^{''}}\in \mathbb{R}^{6N}}$. Subsequently, the dimension of $\mathbf{E}$ is reduced from 6N to M. We set M to the size of action representation $F$ by a linear projection.

\noindent \textbf{Interaction Decoders.} 
Having reasoned about all pairwise interactions using the encoder, we use $\mathbf{E}$ to enrich the action representation, as in \cite{girdhar2019video}.
The pipeline of the decoder is similar to the encoder. Specifically, the features $\mathbf{E}$ from the encoder is projected to key $K^{'}\in \mathbb{R}^{M}$ and value $V^{'}\in \mathbb{R}^{M}$. Different from the initial transformer \cite{vaswani2017attention}, we remove the self attention from the decoder due to action representation features $F \in \mathbb{R}^{M}$ pooled on temporal dimension at backbone features. Instead, we directly map $F$ linearly to $Q^{'} \in \mathbb{R}^{M}$. 
Similarly, we adopt dropout, a feedforward network as well as residual connections, like in Eq.~\ref{equ: ffn}, to learn decoder's output $I \in \mathbb{R}^{M}$.

At last, we stack the multi-head interaction encoders and decoders. The output $I$ is fed to a fully connected layer, and trained to classify actions using standard cross entropy loss. 
During tracking, we backpropagate through all components of the IRN including IU and convolutional positional encoding, as well as the base network. 
The weights of the detector remain frozen.

\section{Experiments and Results}
\label{sec:results}

In this section, we experimentally evaluate the model on two datasets featuring hand-object interactions from egocentric and crowd sourced videos respectively. 

\subsection{Datasets, Evaluation and Implementation Details}
\label{sec: dataset}
\noindent \textbf{EPIC-KITCHENS-100~\cite{Damen2020RESCALING}} is the largest video dataset in egocentric vision. 
We report on the full dataset but the conduct ablation study on a subset of the validation set. The subset was collected by the first participant~(P01) and contains 5,509 / 885 action segments for the training and validation set, respectively.
 We use a fixed random seed and a single run during ablations to ensure results are directly comparable. 

\noindent
\textbf{Something-Else~\cite{materzynska2020something}} is based on~\cite{goyal2017something} proposing a new split for novel verb-noun combinations in the test set. 
Active objects have been densely annotated for training by~\cite{materzynska2020something} -- however hands are not annotated for side L/R and accordingly the active objects are not associated with a hand. 
We use this proposed split without the spatial annotations, instead using automatic, potentially noisy, detections.

\noindent
\textbf{Evaluation} We use evaluation metrics proposed for both dataset - Top-1/5 accuracy for verb, noun and action classes. 
For Something-Else, we use a single output classifier.

\noindent \textbf{Implementation Details.} The hand-object detector \cite{Shan20} aims to find links between hands and interacting objects by optimising an offset vector. It was trained on 100 days dataset that contains both first-person and third-person videos. We use the publicly available trained weights, and detect frame-level hands and objects with the confidence threshold of 0.5.

We use SlowFast \cite{feichtenhofer2019slowfast} R-50 8×8 due to its performance on both datasets. The input videos are 32-frame clips, where we sample $T = 8$ frames with a temporal stride $\tau=8$ for the slow pathway, and 32 frames for the fast pathway. We pool hands and objects features $C=640$ based on the third block of SlowFast as when the layer goes deeper, the features get sparser. We set $N=5120$ used in the encoder and $M=2304$ so that it matches the the globally pooled feature size of the last block of SlowFast. For the interaction unit, 16 heads are used in the multi-head attention and 3-layer encoders and decoders are implemented.

For EPIC-KITCHENS-100, the backbone is pre-trained on the training set and $e()$ and $d()$ are trained from random initialisation. We train for 24 epochs with learning rate 0.001 using SGD with 0.0001 weight decay and 0.9 momentum, the learning rate is decayed by the factor of 10 at epochs 10 and 20. For Something-Else the backbone is pretrained on Kinetics-400 \cite{kay2017kinetics}. Similarly, the network was optimised by SGD with initial learning rate 0.01 dropped at epochs 12 and 18, 0.0001 weight decay, 0.9 momentum for 20 epochs.


\begin{figure}[t]
\centering
\includegraphics[width=1\linewidth]{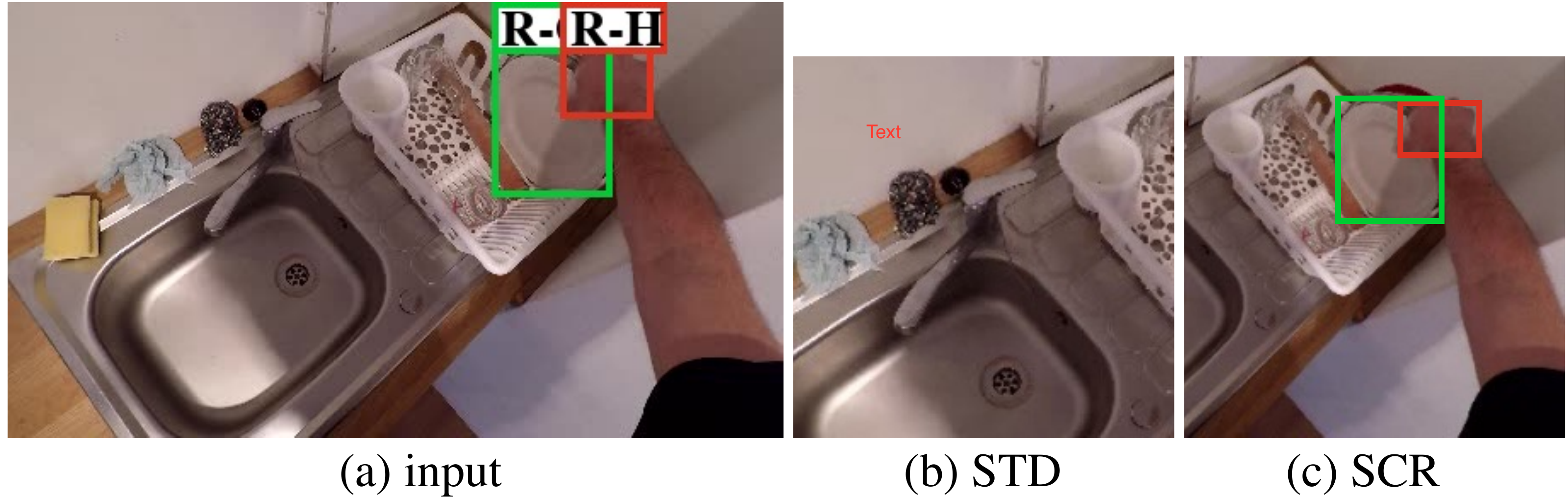}
\vspace*{-6pt}
\caption{Visualisation of STD and SCR for data augmentation. The action is ``take plate''.}
\vspace*{-1pt}
\label{fig: dataAugmentation}
\end{figure}
It is important to note that we changed the random cropping typically used for data augmentation. This is because randomly cropping the image in EPIC-KITCHENS-100 results in hands and objects being frequently cropped out of the main frame. As we show in Fig.~\ref{fig: dataAugmentation}, the standard STD crops the right hand and plate out. This significantly harms the interaction unit. 
In egocentric footage, hands can be in the bottom half of the image or towards a corner. To avoid this while maintaining the power of augmentation, we first randomly scale (\textbf{S}) frames to ${H}'\times {W}'$, then crop (\textbf{C}) frames to ${H}'\times {H}'$ and finally resize (\textbf{R}) frames to the target resolution $224\times 224$. We refer to this augmentation approach as SCR, which we use in all our experiments on EPIC-KITCHENS-100. 
The two data augmentations perform comparably on Something-Else dataset as the action is typically at the centre of the image.
We thus use the standard random cropping data augmentation for this dataset.

\subsection{Something-Else Dataset Results}
\label{subsec: ss}

Tab.~\ref{tbl: mcb} shows the performance of our method outperforms SotA reasoning approaches STIN, STRG and CAF. Specifically, we outperform STIN Combined with I3D, when trained jointly or separately.
Notably, our class-agnostic IRN is superior to methods using object labels among non-ensemble models. 
We also report ensemble methods. 
These metods are not directly comparable but our model remains competitive.

\begin{table}[t]
\centering
\resizebox{0.8\linewidth}{!}{
\begin{tabular}{ccccc}
\hline
Method                                   &Ens. &Obj. & Top-1 & Top-5 \\ \hline
STIN~\cite{materzynska2020something}        &\xmark  &\cmark & 37.2  & 62.4 \\
I3D+STIN~\cite{materzynska2020something}    &\xmark  &\cmark & 48.2  & 72.6 \\
CAF~\cite{radevski2021revisiting} &\xmark  &\cmark & 52.3  & 78.9 \\
STRG~\cite{wang2018videos}       &\xmark  &\xmark & 52.3  & 78.3 \\
IRN (Ours)       &\xmark &\xmark & 52.9  & 80.8 \\ 
\hline
I3D-STIN~\cite{materzynska2020something}    &\cmark  &\cmark & 51.5  & 77.1 \\
STRG-STIN~\cite{materzynska2020something} &\cmark  &\cmark& 56.2  & 81.3 \\ 
CACNF~\cite{radevski2021revisiting} &\cmark  &\cmark & 56.9 & 82.5 \\
\hline
\end{tabular}}\vspace*{-1pt}
\caption{Results on Something-Else Datasets. +: jointly trained. -: trained separately. Ens.: Ensemble. Obj.: use manual object labels. }
\vspace*{-12pt}
\label{tbl: mcb}
\end{table}

Importantly, we showcase that our model particularly benefits from the usage of a SlowFast backbone. We evaluate our interaction units with different backbones in the Tab.~\ref{tbl: bcb}.
We get the best performance when combining SlowFast as a base network with our interaction unit, but also report improvement on the I3D backbone.
\begin{table}[t]
\centering
\resizebox{0.75\linewidth}{!}{
\begin{tabular}{ccccc} \hline
I3D            & SlowFast    & $IRN$         & Top-1 & Top-5 \\ \hline
\checkmark     &             &           & 46.8  & 72.2 \\
\checkmark     &             &\checkmark & 47.5  & 73.8 \\
               &\checkmark   &           & 52.2  & 80.3 \\
               &\checkmark   &\checkmark & 52.9  & 80.8 \\ \hline
\end{tabular}}
\vspace*{-6pt}
\caption{Results of different backbones on Something-Else.}
\label{tbl: bcb}
\end{table}
In the Something-Else dataset, we find that only 39.63\% of clips have both left and right hands, while 50.11\% and 10.26\% of clips have only one hand or no hands respectively. 
When no actor/hand is detected, our interaction unit is likely to struggle.
Despite this, our proposed method can enrich the representation for interaction reasoning.


\subsection{EPIC-KITCHENS-100 Dataset Results}
\label{subsec: epic}

As this dataset is egocentric, i.e. participants are using a wearable camera, more actions involve both hands making it more suitable to assess our proposal.
To manage the size, we conduct an ablation on a selected subset\footnote{We'll make the subset available for direct comparisons}.


\noindent \textbf{Interaction Components.} To evaluate the contribution of the various encoding interactions, we ablate the results by removing one at a time, as well as left/right hand encoders in Tab.~\ref{tbl: IC}. 
We first note that removing right hand interaction encoders (row~6) results in a larger drop than left hand (row~2).
This is anticipated with most participants being right-handed.
Similarly, the largest drop is associated with removing the encoder of $H_R, O_R$, which is critical for one-handed interactions (row~7) followed by $H_R, H_L$ (row~9) which is critical for hand-only interactions (e.g. wash hands).
The encoder with the least contribution is that for $H_R, O_L$ (row~8), as it is probably compensated by the other pairwise encoders.

\begin{figure*}[t!]
\centering
\includegraphics[width=1\linewidth]{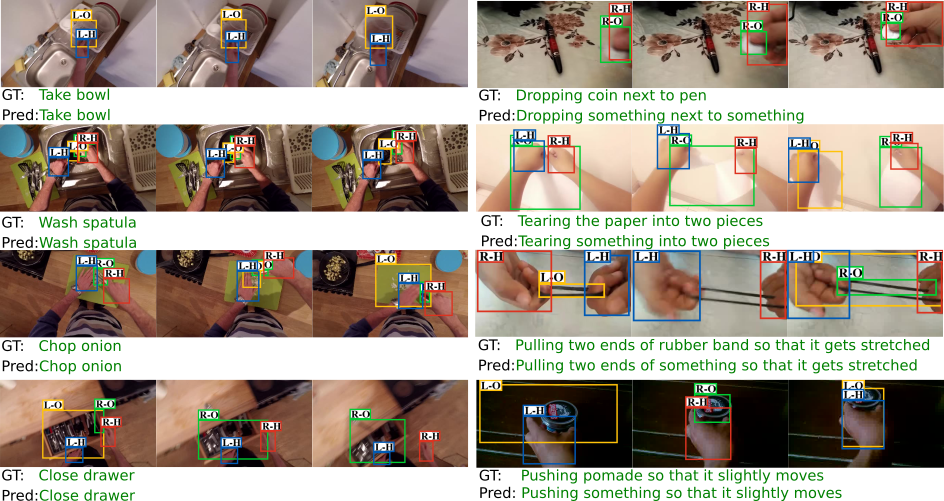}
\caption{Qualitative results of correctly-recognised interactions from EPIC-KITCHENS-100 (col 1) and Something-Else~(col~2). $L/ R$ indicate the side and $H/O$ denote hands and objects. $GT$ and $Pred$ are Ground Truth and Prediction.}
\vspace*{-6pt}
\label{fig:SE}
\end{figure*}

\noindent \textbf{Trajectory.} We evaluate the importance of using trajectories. Previous work~\cite{girdhar2019video} only uses one actor (person) of the middle frame ($middle$) along with video action representation and a recent work~\cite{pan2020actor} detects persons in the middle frame and then duplicate these detections across time ($duplicate$).
We compare these options to our proposed approach that encodes object trajectories as well as two-handed interactions. 
The best performance is achieved when using the complete trajectory of detections, including both hands and objects. For more details, please refer to the supplementary.

\begin{table}[t!]
\footnotesize
\begin{minipage}[t]{1\linewidth}
\begin{center}
\resizebox{\linewidth}{!}{
\begin{tabular}{ccccccc}
\hline
\tiny{$(H_L,O_L)$}  & \tiny{$(H_L,O_R)$} & \tiny{$(H_L, H_R)$}  &\tiny{$(H_R, O_R)$}   & \tiny{$(H_R, O_L)$}  & \tiny{$(H_R, H_L)$}    & Top-1  \\ \hline
\xmark&\xmark &\xmark &\xmark  &\xmark &\xmark   &42.37  \\
\xmark&\xmark &\xmark &\checkmark  &\checkmark &\checkmark   &43.28  \\

\xmark &\checkmark  &\checkmark &\checkmark  &\checkmark&\checkmark             &42.60 \\
\checkmark&\xmark  &\checkmark &\checkmark  &\checkmark&\checkmark              &43.28 \\
\checkmark&\checkmark  &\xmark &\checkmark  &\checkmark&\checkmark              &42.94\\
\checkmark&\checkmark  &\checkmark &\xmark   &\xmark  &\xmark   &42.82  \\

\checkmark&\checkmark  &\checkmark &\xmark  &\checkmark&\checkmark      &41.47 \\
\checkmark&\checkmark  &\checkmark &\checkmark  &\xmark &\checkmark          &44.07  \\
\checkmark&\checkmark  &\checkmark &\checkmark  &\checkmark &\xmark         &41.69  \\
\checkmark&\checkmark  &\checkmark &\checkmark  &\checkmark &\checkmark &\textbf{44.52}  \\
\hline
\end{tabular}
}
\end{center}
\vspace*{-1pt}
\caption{Ablation study for Interaction Components.}
\vspace*{6pt}
\label{tbl: IC}
\end{minipage}
\vspace*{-12pt}
\end{table}
\begin{table}[t]
\begin{minipage}[t]{1\linewidth}
\footnotesize
\begin{center}
\setlength{\tabcolsep}{2mm}{
\begin{tabular}{cccccc}
\hline
 Det.       &$H_R$ &Act. Rep. &$H_L$      &Objects       & Top-1  \\ \hline
 middle~\cite{girdhar2019video}     &\checkmark  &\checkmark &  &    &43.05  \\
 duplicate~\cite{pan2020actor}   &\checkmark &\checkmark &  &       &44.07  \\
  trajectory &\checkmark   &\checkmark & &                          &42.37  \\
  trajectory &\checkmark   &\checkmark & &\checkmark                 &43.28 \\
 trajectory &\checkmark &\checkmark &\checkmark &\checkmark   &\textbf{44.52}   \\\hline
\end{tabular}
}
\end{center}
\vspace*{-1pt}
\caption{Comparison to prior works without trajectory representations as well as active object trajectories.}
\vspace*{-1pt}
\label{tbl: traj}
\end{minipage} 
\end{table}
\begin{table}[t]
\begin{minipage}[t]{0.25\linewidth}
\footnotesize
\begin{center}
\begin{tabular}{ll}
\hline
 SPE          & Top-1   \\ \hline
 none            &43.39   \\
 concat        &42.71  \\
 sum         &\textbf{44.52}  \\ \hline
\end{tabular}
\end{center}
\caption{Ablation study for Spatial Position Encoding. }
\label{tbl: inter sep}
\end{minipage}
\hspace{0.05\linewidth}
\begin{minipage}[t]{0.25\linewidth}
\footnotesize
\begin{center}
\begin{tabular}{ll}
\hline
 Act.Rep.       &Top-1  \\ \hline
 none        &31.98  \\
 concat   &43.73 \\
 decoder  &\textbf{44.52}  \\ \hline
\end{tabular}
\end{center}
\caption{Ablation study for Action Representation.}
\label{tbl: context}
\end{minipage}
\vspace*{-1pt}
\hspace{0.05\linewidth}
\begin{minipage}[t]{0.25\linewidth}
\footnotesize
\begin{center}
\begin{tabular}{ll}
\hline
Det.Rep       &Top-1  \\ \hline
 MLP   &43.95 \\
 CNN  &\textbf{44.52}  \\ \hline
\end{tabular}
\end{center}
\caption{Ablation study for Detection Representation.}
\label{tbl: vfe}
\end{minipage}
\end{table}


\noindent \textbf{Spatial Positional Encoding.} We adopt two ways to fuse positional information with visual features of hands or objects. As we expected in Tab.~\ref{tbl: inter sep}, spatial position encoding $SPE$ improves the performance. Compared to the network without $SPE$ ($none$ in Tab.~\ref{tbl: inter sep}) or concatenations of $SPE$. This is due to the sparsity of positional features. Concatenating lots of zeros to the visual features may introduce noise. Our proposed approach to $sum$ the positional encoding yields the best action performance.

\noindent \textbf{Action Representation.} In this ablation study, we evaluate the importance of globally pooled action representation. First, We remove decoders
and train the network with only interaction features, i.e. $none$. Tab.~\ref{tbl: context} shows that results drop significantly if the network does not use the globally-pooled action representation. We also concatenate the hand-object interaction features with action representation, in a late-fusion fashion.
What stands out in the Tab.~\ref{tbl: context} is that the Top-1 action result of the decoder surpasses concatenation. 

\noindent \textbf{Detection Representation.} 
In all other experiments, we have used RoI pooling to get feature representations of our hands' and objects' detections. 
Instead, inspired by popular MLPs~\cite{tolstikhin2021mlp}, we also explore using MLP projection directly from image patches. We crop the input image given bounding boxes $b$ and resize to a fixed size by bilinear interpolation. The fixed size patches are flatten, and projected using an MLP. 
Hence, representations of any detections can be learned by a parameter-shared MLP. Notably, these features only contain the spatial information, without any motion cues as they are projected per image. We compare performances of MLPs and pooled 3D-CNN as the features extractors. Tab.~\ref{tbl: vfe} shows that the accuracy of 3D-CNN pooling is superior to MLP projection, indicating that temporal features are significant for video-based hand-to-object interactions.

\noindent \textbf{Detections Feature Representation.} 
In all other experiments, we have used RoI pooling to get feature representations of our detections. As presented in Sec~\ref{sec:method}, we also explore using MLP projection.
Inspired by popular MLPs \cite{tolstikhin2021mlp}, we compare performances of MLPs and ConvNets as to get the visual features of our detections. Tab.~\ref{tbl: vfe} shows that MLP has better modelling capabilities than ConvNets only for verb recognition. However, ConvNets are more advantageous for noun recognition and their accuracy is far better than MLPs by about 2.7\%. Hence, ConvNets are more accurate in action recognition.
All previous and subsequent results thus use 3D ConvNet RoI pooling to represent the visual features of hand and object detections.
\begin{table}[t]
\centering
\resizebox{0.8\linewidth}{!}{%
\begin{tabular}{ccccccccc}
\hline
  &        & \multicolumn{3}{l}{Top-1 Accuracy (\%)}\\ \cline{3-5}
Method      &DataAug.      & Verb   & Noun   & Act. \\ \hline
Chance~\cite{Damen2020RESCALING}  &STD &10.42 &1.70 &0.51\\
IRN (Ours)          &STD &60.94   &43.97   &31.97  \\
TSN~\cite{Damen2020RESCALING}  &STD &60.18 &46.03 &33.19\\
SlowFast \cite{Damen2020RESCALING}     &SCR &63.64   &48.58   &36.76  \\
IRN (Ours)          &SCR &\textbf{63.68}   &\textbf{48.94}   &\textbf{37.11}  \\ \hline
\end{tabular}}
\caption{Quantitative results on Validation set. STD and SCR denote standard and our data augmentation strategies.}
\vspace*{-1pt}
\label{tbl: da}
\end{table}

We next report results on the validation set of EPIC-KITCHENS-100. In Tab.~\ref{tbl: da}, 
our method $IRN$ improves over published baselines. Adding the interaction unit to SlowFast improves results for verb, noun and action top-1 accuracy by 0.04\%, 0.36\% and 0.35\%, respectively. 
We explain the data augmentation strategies in supplementary, where STD performs random cropping often cropping the hands out of the image. 
Comparing row 2 and row 5, we demonstrate that the random cropping harms the performance of IRN.


\subsection{Qualitative Results}
\label{sec:discussion}

\begin{figure*}[t]
\centering
\label{fig: SE}
\includegraphics[width=1\linewidth]{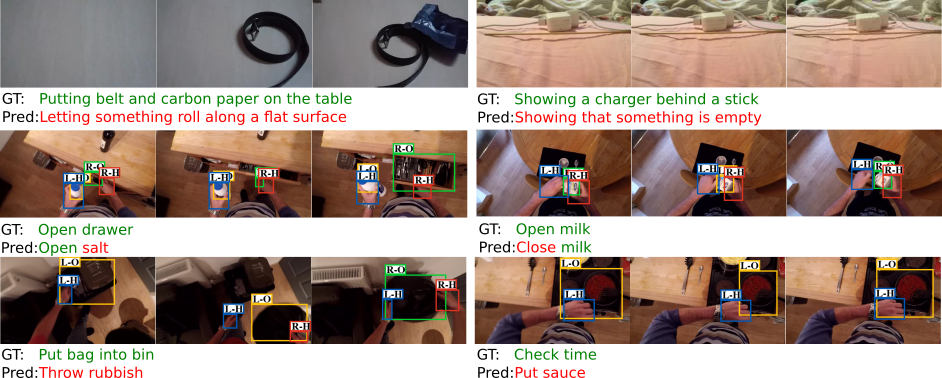}
\vspace*{-8pt}
\caption{Failure cases of Something-Else (row 1) and EPIC-KITCHENS-100 (rows 2, 3).}
\vspace*{-6pt}
\label{fig: DI}
\end{figure*}

We demonstrate the results on clips containing only one hand, or both hands in Fig.~\ref{fig:SE}. It illustrates correct action recognition examples from both datasets. 
It is also important to highlight that our method recovers from partial failures in automatic detections of hands and objects, as in rows~3 and~4. IRN is robust to two types of detection errors:
(a) incorrect active objects detection; (b) incorrect side of hand and objects. Specifically, row 3 illustrates ``chop onion'' and ``pull two ends of something so that it gets stretched''. The onion is not detected in several frames due to occlusion. Similarly. the knife and rubber are missed in several frames along the trajectory. For row 4 (col 1) the drawer switches between being a left-object or a right object, while in row 4 (col 2) the hand sides are swapped in various frames. Both errors (a) and (b) have little impact on our method, due to our usage of trajectories and the attention mechanism that selects the relevant hand and object representations to reason about interactions. 

Moreover, we show failure cases in Fig.~\ref{fig: DI}. The main reason for failure is  undetected or unobserved hands throughout the videos (row 1). In both examples, the hands are not visible throughout. Row 2 (col 1) shows that our focus on both hands might result in detecting another concurrent action, paying more attention to the object in the other hand (salt).
In row 2 (col 2) the method fails to recognise the interaction with 
small movement.
In row 3, the less frequent action of replacing the bin bag is incorrectly mistaken as the frequent throwing action. 
A clear limitation of our approach is evident in row 3 (col 2) where the person is checking the time on their hand watch. Evidently, the watch is never recognised as the interacting object, as it is always part of the hand detection.

\section{Conclusion}
\label{sec:conclusion}
In this paper, we present a framework for hand-object interaction reasoning that separately attends to actors (hands) and interacting objects, through encoders, as well as action representation which includes contextual information through a decoder. We present results on two hand-object interaction datasets, demonstrating generality and competitive performance, with an ablation study. 

\small\noindent\textbf{Acknowledgements.} This work used public datasets and is supported by EPSRC UMPIRE (EP/T004991/1).

\nocite{*} 
\bibliographystyle{ieee_fullname}
\bibliography{main} 

\begin{thebibliography}{10}\itemsep=-1pt

\bibitem{Baradel_2018_ECCV}
Fabien Baradel, Natalia Neverova, Christian Wolf, Julien Mille, and Greg Mori.
\newblock Object level visual reasoning in videos.
\newblock In {\em ECCV}, 2018.

\bibitem{beltagy2020longformer}
Iz Beltagy, Matthew~E Peters, and Arman Cohan.
\newblock Longformer: The long-document transformer.
\newblock {\em arXiv preprint arXiv:2004.05150}, 2020.

\bibitem{Bertasius2021}
Gedas Bertasius, Heng Wang, and Lorenzo Torresani.
\newblock Is space-time attention all you need for video understanding?
\newblock In {\em ICML}, 2021.

\bibitem{carreira2017quo}
Joao Carreira and Andrew Zisserman.
\newblock Quo vadis, action recognition? a new model and the kinetics dataset.
\newblock In {\em CVPR}, pages 6299--6308, 2017.

\bibitem{chao2018learning}
Yu-Wei Chao, Yunfan Liu, Xieyang Liu, Huayi Zeng, and Jia Deng.
\newblock Learning to detect human-object interactions.
\newblock In {\em WACV}, pages 381--389, 2018.

\bibitem{Damen2020RESCALING}
Dima Damen, Hazel Doughty, Giovanni~Maria Farinella, , Antonino Furnari, Jian
  Ma, Evangelos Kazakos, Davide Moltisanti, Jonathan Munro, Toby Perrett, Will
  Price, and Michael Wray.
\newblock Rescaling egocentric vision.
\newblock {\em CoRR}, abs/2006.13256, 2020.

\bibitem{Damen2018EPICKITCHENS}
Dima Damen, Hazel Doughty, Giovanni~Maria Farinella, Sanja Fidler, Antonino
  Furnari, Evangelos Kazakos, Davide Moltisanti, Jonathan Munro, Toby Perrett,
  Will Price, and Michael Wray.
\newblock Scaling egocentric vision: The epic-kitchens dataset.
\newblock In {\em ECCV}, 2018.

\bibitem{donahue2015long}
Jeffrey Donahue, Lisa Anne~Hendricks, Sergio Guadarrama, Marcus Rohrbach,
  Subhashini Venugopalan, Kate Saenko, and Trevor Darrell.
\newblock Long-term recurrent convolutional networks for visual recognition and
  description.
\newblock In {\em CVPR}, pages 2625--2634, 2015.

\bibitem{fan2020pyslowfast}
Haoqi Fan, Yanghao Li, Bo Xiong, Wan-Yen Lo, and Christoph Feichtenhofer.
\newblock Pyslowfast.
\newblock \url{https://github.com/facebookresearch/slowfast}, 2020.

\bibitem{Feichtenhofer_2020_CVPR}
Christoph Feichtenhofer.
\newblock X3d: Expanding architectures for efficient video recognition.
\newblock In {\em CVPR}, June 2020.

\bibitem{feichtenhofer2019slowfast}
Christoph Feichtenhofer, Haoqi Fan, Jitendra Malik, and Kaiming He.
\newblock Slowfast networks for video recognition.
\newblock In {\em ICCV}, pages 6202--6211, 2019.

\bibitem{feichtenhofer2016convolutional}
Christoph Feichtenhofer, Axel Pinz, and Andrew Zisserman.
\newblock Convolutional two-stream network fusion for video action recognition.
\newblock In {\em CVPR}, pages 1933--1941, 2016.

\bibitem{gao2020drg}
Chen Gao, Jiarui Xu, Yuliang Zou, and Jia-Bin Huang.
\newblock Drg: Dual relation graph for human-object interaction detection.
\newblock In {\em ECCV}, pages 696--712, 2020.

\bibitem{gao2018ican}
Chen Gao, Yuliang Zou, and Jia-Bin Huang.
\newblock ican: Instance-centric attention network for human-object interaction
  detection.
\newblock In {\em BMVC}, 2018.

\bibitem{girdhar2019video}
Rohit Girdhar, Joao Carreira, Carl Doersch, and Andrew Zisserman.
\newblock Video action transformer network.
\newblock In {\em CVPR}, pages 244--253, 2019.

\bibitem{gkioxari2018detecting}
Georgia Gkioxari, Ross Girshick, Piotr Doll{\'a}r, and Kaiming He.
\newblock Detecting and recognizing human-object interactions.
\newblock In {\em CVPR}, pages 8359--8367, 2018.

\bibitem{goyal2017something}
Raghav Goyal, Samira Ebrahimi~Kahou, Vincent Michalski, Joanna Materzynska,
  Susanne Westphal, Heuna Kim, Valentin Haenel, Ingo Fruend, Peter Yianilos,
  Moritz Mueller-Freitag, et~al.
\newblock The" something something" video database for learning and evaluating
  visual common sense.
\newblock In {\em ICCV}, pages 5842--5850, 2017.

\bibitem{Gu_2018_CVPR}
Chunhui Gu, Chen Sun, David~A. Ross, Carl Vondrick, Caroline Pantofaru, Yeqing
  Li, Sudheendra Vijayanarasimhan, George Toderici, Susanna Ricco, Rahul
  Sukthankar, Cordelia Schmid, and Jitendra Malik.
\newblock Ava: A video dataset of spatio-temporally localized atomic visual
  actions.
\newblock In {\em CVPR}, June 2018.

\bibitem{he2017mask}
Kaiming He, Georgia Gkioxari, Piotr Doll{\'a}r, and Ross Girshick.
\newblock Mask r-cnn.
\newblock In {\em ICCV}, pages 2961--2969, 2017.

\bibitem{islam2020much}
Md~Amirul Islam, Sen Jia, and Neil~DB Bruce.
\newblock How much position information do convolutional neural networks
  encode?
\newblock In {\em ICLR}, 2020.

\bibitem{ji20123d}
Shuiwang Ji, Wei Xu, Ming Yang, and Kai Yu.
\newblock 3d convolutional neural networks for human action recognition.
\newblock {\em TPAMI}, 35(1):221--231, 2012.

\bibitem{kay2017kinetics}
Will Kay, Joao Carreira, Karen Simonyan, Brian Zhang, Chloe Hillier, Sudheendra
  Vijayanarasimhan, Fabio Viola, Tim Green, Trevor Back, Paul Natsev, et~al.
\newblock The kinetics human action video dataset.
\newblock {\em arXiv preprint arXiv:1705.06950}, 2017.

\bibitem{li2019collaborative}
Chao Li, Qiaoyong Zhong, Di Xie, and Shiliang Pu.
\newblock Collaborative spatiotemporal feature learning for video action
  recognition.
\newblock In {\em CVPR}, pages 7872--7881, 2019.

\bibitem{li2019transferable}
Yong-Lu Li, Siyuan Zhou, Xijie Huang, Liang Xu, Ze Ma, Hao-Shu Fang, Yanfeng
  Wang, and Cewu Lu.
\newblock Transferable interactiveness knowledge for human-object interaction
  detection.
\newblock In {\em CVPR}, pages 3585--3594, 2019.

\bibitem{lin2019tsm}
Ji Lin, Chuang Gan, and Song Han.
\newblock Tsm: Temporal shift module for efficient video understanding.
\newblock In {\em IOC]}, pages 7083--7093, 2019.

\bibitem{materzynska2020something}
Joanna Materzynska, Tete Xiao, Roei Herzig, Huijuan Xu, Xiaolong Wang, and
  Trevor Darrell.
\newblock Something-else: Compositional action recognition with
  spatial-temporal interaction networks.
\newblock In {\em CVPR}, pages 1049--1059, 2020.

\bibitem{neimark2021video}
Daniel Neimark, Omri Bar, Maya Zohar, and Dotan Asselmann.
\newblock Video transformer network.
\newblock {\em arXiv preprint arXiv:2102.00719}, 2021.

\bibitem{pan2021actor}
Junting Pan, Siyu Chen, Mike~Zheng Shou, Yu Liu, Jing Shao, and Hongsheng Li.
\newblock Actor-context-actor relation network for spatio-temporal action
  localization.
\newblock In {\em CVPR}, pages 464--474, 2021.

\bibitem{pan2020actor}
Junting Pan, Siyu Chen, Zheng Shou, Jing Shao, and Hongsheng Li.
\newblock Actor-context-actor relation network for spatio-temporal action
  localization.
\newblock 2020.

\bibitem{radevski2021revisiting}
Gorjan Radevski, Marie-Francine Moens, and Tinne Tuytelaars.
\newblock Revisiting spatio-temporal layouts for compositional action
  recognition.
\newblock {\em arXiv preprint arXiv:2111.01936}, 2021.

\bibitem{ren2015faster}
Shaoqing Ren, Kaiming He, Ross Girshick, and Jian Sun.
\newblock Faster r-cnn: Towards real-time object detection with region proposal
  networks.
\newblock In {\em NeurIPS}, pages 91--99, 2015.

\bibitem{Shan20}
Dandan Shan, Jiaqi Geng, Michelle Shu, and David Fouhey.
\newblock Understanding human hands in contact at internet scale.
\newblock In {\em CVPR}, 2020.

\bibitem{simonyan2014two}
Karen Simonyan and Andrew Zisserman.
\newblock Two-stream convolutional networks for action recognition in videos.
\newblock In {\em NeurIPS}, 2014.

\bibitem{soomro2014action}
Khurram Soomro and Amir~R Zamir.
\newblock Action recognition in realistic sports videos.
\newblock In {\em the Computer Vision in Sports}, pages 181--208. 2014.

\bibitem{sun2018actor}
Chen Sun, Abhinav Shrivastava, Carl Vondrick, Kevin Murphy, Rahul Sukthankar,
  and Cordelia Schmid.
\newblock Actor-centric relation network.
\newblock In {\em ECCV}, pages 318--334, 2018.

\bibitem{tolstikhin2021mlp}
Ilya Tolstikhin, Neil Houlsby, Alexander Kolesnikov, Lucas Beyer, Xiaohua Zhai,
  Thomas Unterthiner, Jessica Yung, Daniel Keysers, Jakob Uszkoreit, Mario
  Lucic, et~al.
\newblock Mlp-mixer: An all-mlp architecture for vision.
\newblock {\em arXiv preprint arXiv:2105.01601}, 2021.

\bibitem{tran2015learning}
Du Tran, Lubomir Bourdev, Rob Fergus, Lorenzo Torresani, and Manohar Paluri.
\newblock Learning spatiotemporal features with 3d convolutional networks.
\newblock In {\em ICCV}, pages 4489--4497, 2015.

\bibitem{vaswani2017attention}
A Vaswani, Noam Shazeer, Niki Parmar, Jakob Uszkoreit, Llion Jones, Aidan~N
  Gomez, Lukasz Kaiser, and Illia Polosukhin.
\newblock Attention is all you need.
\newblock In {\em NeurIPS}, 2017.

\bibitem{wan2019pose}
Bo Wan, Desen Zhou, Yongfei Liu, Rongjie Li, and Xuming He.
\newblock Pose-aware multi-level feature network for human object interaction
  detection.
\newblock In {\em ICCV}, pages 9469--9478, 2019.

\bibitem{wang2016temporal}
Limin Wang, Yuanjun Xiong, Zhe Wang, Yu Qiao, Dahua Lin, Xiaoou Tang, and Luc
  Van~Gool.
\newblock Temporal segment networks: Towards good practices for deep action
  recognition.
\newblock In {\em ECCV}, pages 20--36, 2016.

\bibitem{wang2018non}
Xiaolong Wang, Ross Girshick, Abhinav Gupta, and Kaiming He.
\newblock Non-local neural networks.
\newblock In {\em CVPR}, pages 7794--7803, 2018.

\bibitem{wang2018videos}
Xiaolong Wang and Abhinav Gupta.
\newblock Videos as space-time region graphs.
\newblock In {\em ECCV}, pages 399--417, 2018.

\bibitem{lfb2019}
Chao-Yuan Wu, Christoph Feichtenhofer, Haoqi Fan, Kaiming He, Philipp
  Kr\"{a}henb\"{u}hl, and Ross Girshick.
\newblock {Long-Term Feature Banks for Detailed Video Understanding}.
\newblock In {\em CVPR}, 2019.

\bibitem{yue2015beyond}
Joe Yue-Hei~Ng, Matthew Hausknecht, Sudheendra Vijayanarasimhan, Oriol Vinyals,
  Rajat Monga, and George Toderici.
\newblock Beyond short snippets: Deep networks for video classification.
\newblock In {\em CVPR}, pages 4694--4702, 2015.

\bibitem{zhou2018temporal}
Bolei Zhou, Alex Andonian, Aude Oliva, and Antonio Torralba.
\newblock Temporal relational reasoning in videos.
\newblock In {\em ECCV}, pages 803--818, 2018.

\bibitem{zhou2019relation}
Penghao Zhou and Mingmin Chi.
\newblock Relation parsing neural network for human-object interaction
  detection.
\newblock In {\em ICCV}, pages 843--851, 2019.

\end{thebibliography}

\end{document}